%% file: main.tex
\documentclass[letterpaper, 10 pt, conference]{ieeeconf} 
\IEEEoverridecommandlockouts

\IEEEoverridecommandlockouts                              

\usepackage{amsmath,mleftright,mathtools}
\usepackage{amssymb}
\usepackage{lipsum}
\usepackage[pdftex, pdfstartview={FitV}, pdfpagelayout={TwoColumnLeft},bookmarksopen=true,plainpages = false, colorlinks=true, linkcolor=black, citecolor = black, urlcolor = blue,filecolor=black , pagebackref=false,hypertexnames=false, plainpages=false, pdfpagelabels ]{hyperref}

\usepackage{balance}
\usepackage{xargs}
\usepackage{graphicx}
\usepackage[labelformat=simple]{subcaption}
\DeclareCaptionLabelSeparator{periodspace}{.\quad}
\usepackage[sort,compress]{cite}

\usepackage{afterpage}
\usepackage{tabularx,booktabs}
\usepackage{multirow}
\usepackage{stackengine}
\usepackage{bm,etoolbox}
\usepackage[linesnumbered,ruled,vlined]{algorithm2e}
\usepackage[pdftex,dvipsnames]{xcolor}
\usepackage{rotating}
\usepackage[makeroom]{cancel}

\newcolumntype{C}{>{\centering\arraybackslash}X} 
\newcolumntype{L}{>{\raggedleft\arraybackslash}X} 

\usepackage{dblfloatfix}

\usepackage{todonotes}
\newcommandx{\unsure}[2][1=]{\todo[linecolor=red,backgroundcolor=red!25,bordercolor=red,#1]{#2}}

\newcommand{\highlight}[1]{{\color{black}#1}}

\DeclareGraphicsExtensions{.pdf,.png}
\graphicspath{ {./figs/} }
\DeclarePairedDelimiterX{\infdivx}[2]{(}{)}{%
  #1\;\delimsize\|\;#2%
}

\SetCommentSty{mycommfont}
\include{notation}
\title{\LARGE \bf
GAPSLAM: Blending Gaussian Approximation and Particle Filters for Real-Time Non-Gaussian SLAM
}

\author{Qiangqiang Huang$^{1}$ and John J. Leonard$^{1}$
\thanks{$^{1}$Computer Science and Artificial Intelligence Lab,
Massachusetts Institute of Technology, Cambridge, MA 02139, USA{\tt\small \{hqq, jleonard\}@mit.edu}
}%
\thanks{
This work was supported by ONR grant N00014-18-1-2832, ONR Neuroautonomy MURI grant N00014-19-1-2571 and the MIT Portugal Program. The project page is available at \url{https://github.com/doublestrong/gapslam}.
}%
}

\begin{document}


\maketitle
\thispagestyle{empty}
\pagestyle{empty}

\input{abstract}
\input{introduction}
\input{related_work}
\input{methods}
\input{results}
\input{conclusion}

\bibliographystyle{IEEEtran}
\bibliography{ref.bib}
\end{document}

%% file: notation.tex
\def \measVal{z}
\def \measVals{\mathbf{z}}

\def \odomVal{o}
\def \odomVals{\mathbf{o}}

\def \daVal{d}
\def \daVals{\mathbf{d}}

\def \idx{\mathcal{V}}

\def \lmkVar{L}
\def \lmkVal{l}
\def \rbtVar{X}
\def \rbtVal{x}
\def \lmkVars{\mathbf{\lmkVar}}
\def \rbtVars{\mathbf{\rbtVar}}

\DeclareMathOperator{\SE}{SE}

\def \potential{\psi}

\DeclareMathOperator*{\argmax}{argmax}

\def \uct{\mathcal{N}} 
\def \gaus{\mathcal{G}} 

%% file: abstract.tex
\begin{abstract}
Inferring the posterior distribution in SLAM is critical for evaluating the uncertainty in localization and mapping, as well as supporting subsequent planning tasks aiming to reduce uncertainty for safe navigation. However, real-time full posterior inference techniques, such as Gaussian approximation and particle filters, either lack expressiveness for representing non-Gaussian posteriors or suffer from performance degeneracy when estimating high-dimensional posteriors. Inspired by the complementary strengths of Gaussian approximation and particle filters\textemdash scalability and non-Gaussian estimation, respectively\textemdash we blend these two approaches to infer marginal posteriors in SLAM. Specifically, Gaussian approximation provides robot pose distributions on which particle filters are conditioned to sample landmark marginals. In return, the maximum a posteriori point among these samples can be used to reset linearization points in the nonlinear optimization solver of the Gaussian approximation, facilitating the pursuit of global optima. We demonstrate the scalability, generalizability, and accuracy of our algorithm for real-time full posterior inference on realworld range-only SLAM and object-based bearing-only SLAM datasets.
\end{abstract}

%% file: introduction.tex
\section{Introduction}
Simultaneous localization and mapping (SLAM) is a foundational technique for autonomous driving, AR/VR, and robotic navigation. While the primary goal of SLAM is to find a point estimate of robot paths and maps, inferring the posterior density encountered in SLAM is also essential. This is because full posterior inference reveals how uncertain the robot path and map could be, and supports the robot in planning how to reduce the uncertainty for safe navigation \cite{Rosen2021Advances}.

Standard \emph{real-time} methods for full posterior inference in SLAM research either compute a Gaussian approximation, centered on the point estimate, to the posterior \cite{dellaert2012factor}, or represent the posterior using samples in the framework of Rao-Blackwellized particle filtering (RBPF) \cite{montemerlo2003fastslam}. Constructing the Gaussian approximation enjoys great advantages in scalability for tackling high-dimensional posteriors, owing to efficient nonlinear optimization solvers for SLAM \cite{kaess2012isam2, kummerle2011g}. However, the Gaussian approximation is inherently incapable of describing highly non-Gaussian/multi-modal posteriors, which often appear in realworld SLAM problems due to nonlinear measurement models. While the RBPF can capture non-Gaussian features of the posterior via samples, it suffers from scalability issues incurred by particle degeneracy and depletion \cite{arulampalam2002tutorial}.

\begin{figure}[t]
    \centering
    \includegraphics[width=\linewidth]{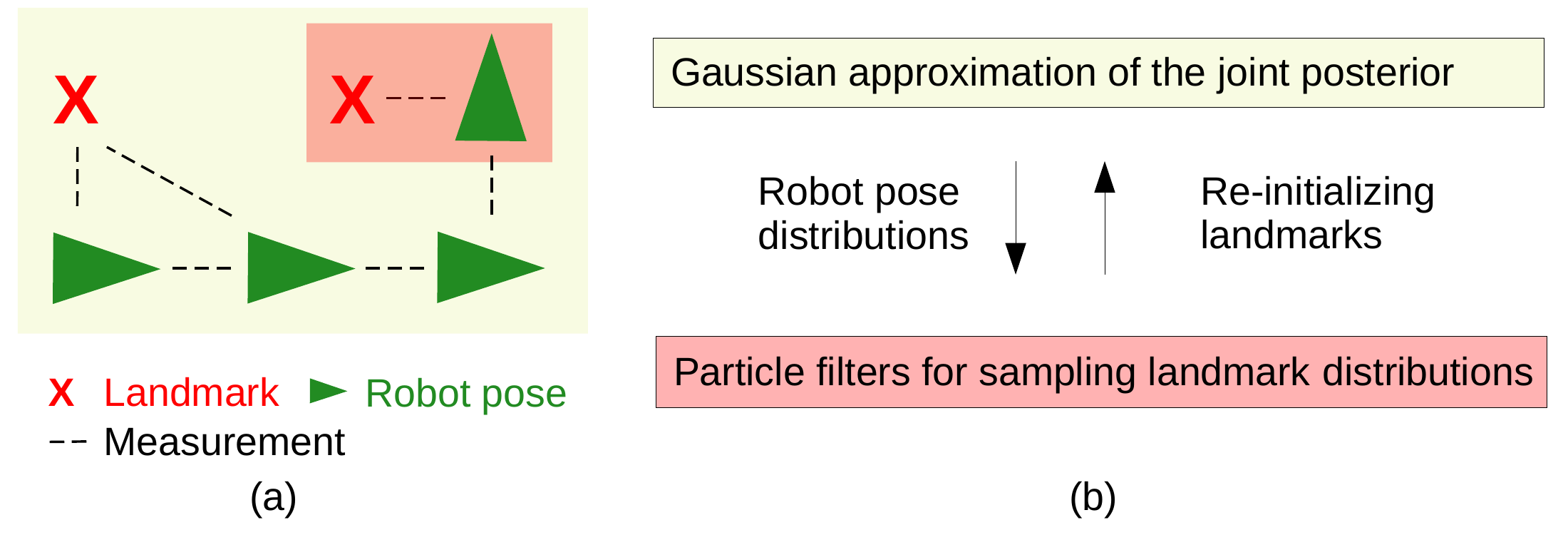}
    \caption{Illustration of our method for inferring robot pose and landmark distributions: (a) a SLAM example, where the robot moves along poses in green and makes measurements to landmarks in red, and (b) our method, which blends Gaussain approximation in yellow and particle filters in pink. The Gaussian approximation, centered on the maximum a posteriori (MAP) estimate, provides distributions of robot poses on which the particle filters are conditioned to draw samples that represent landmark distributions. If a sample attains a higher probability than the MAP estimate when evaluating the posterior, landmarks in the Gaussian solver will be re-initialized by that sample.}
    \label{fig:my_label}
	\vspace{-15pt}
\end{figure}

By blending advantages of the \textbf{G}aussian \textbf{A}pproximation and \textbf{P}article filters on scalability and non-Gaussian estimation, respectively, we present a novel algorithm, \textbf{GAP}SLAM, to infer marginal posteriors encountered in SLAM. The contributions of the paper include:
\begin{enumerate}
\item An adaptive modeling strategy whereby only marginals with great uncertainty are sampled by particle filters, while others are represented by the Gaussian approximation.
\item An uncertainty-aware re-initialization technique that leverages particle filters to reset linearization points in the nonlinear optimization solver.
\item Range-only and object-based bearing-only SLAM experiments that demonstrate the scalability, generalizability, and accuracy of GAPSLAM, as well as its ability to precisely describe the evolution of non-Gaussian posteriors in real-time.
\end{enumerate}

%% file: related_work.tex
\section{Related Work}
We briefly review some work on full posterior inference for SLAM with a focus on non-Gaussian representations of the posterior. Note that the non-Gaussian representation can be parametric models (e.g., sum of Gaussians) or non-parametric (i.e., samples).

The framework of RBPF has been leveraged in a big class of SLAM algorithms (e.g., FastSLAM) \cite{montemerlo2003fastslam, blanco2008pure, blanco2008efficient}, where the posterior of robot poses is represented by a set of particles, and each particle is attached with parametric models \cite{montemerlo2003fastslam, blanco2008efficient} or samples \cite{blanco2008pure} of the conditional of the map. Although our work exploits the same conditional independence relation as FastSLAM, i.e., landmarks are conditionally independent given robot poses, our work swaps representations of the robot path and map in RBPF, resulting in a parametric (Gaussian) model to describe the belief about robot poses and a set of particle filters to independently draw samples from the posterior of each landmark. Thus the issue of losing diversity in robot path particles no longer exists, warranting the scalability of our work. We will compare to \cite{blanco2008efficient} in our experiments section. mm-iSAM \cite{fourie2016nonparametric} and NF-iSAM \cite{huang2021online} are recent non-Gaussian inference algorithms that leverage the Bayes tree algorithm \cite{kaess2012isam2} to exploit more conditional independence relations. As solvers for general factor graphs, they can tackle both nonlinear measurement models and multi-modal data association, while our work is only focused on nonlinear measurement models. Another class of approaches purely relies on Monte Carlo techniques to directly draw samples from the joint posterior \cite{torma2010markov,shariff15aistats, huang2021reference}. These methods, in general, do not suit real-time online applications due to the exorbitant computational cost but can serve as reference solutions to full posterior inference.

Our work is inspired by some works \cite{davison2003real,blanco2008efficient,blanco2008pure} dedicated to bearing-only (point-object-based) or range-only SLAM. Existing approaches for these two problems can be categorized as batch optimizations \cite{newman2003pure}, delayed initializations of landmarks \cite{lemaire2005practical,davison2003real}, and undelayed initializations of landmarks \cite{kwok02iros,sola05iros,blanco2008efficient,blanco2008pure}. Our work falls into the undelayed category. Our experiments show that we can solve both of these problems in a unified framework and demonstrate the evolution of landmark posteriors since time step zero. Furthermore, we clarify that, while our object-based SLAM system requires similar input (RGB images and object detections) as the systems in \cite{nicholson2018quadricslam, yang2019cubeslam, ok2019robust, rubino20173d}, we do not estimate occupied areas of objects and focus solely on inferring how distributions of object locations evolve.

%% file: methods.tex
\section{Proposed Approach}
\highlight{We focus on landmark-based SLAM}. Let $\rbtVars_t:=(\rbtVar_0,\rbtVar_1,\ldots,\rbtVar_t)$ be robot pose variables up to time step $t$ where $\rbtVar_i \in \SE(d)$. Landmark variables are denoted by $\lmkVars_n:=(\lmkVar_1,\lmkVar_2,\dots,\lmkVar_n)$ \highlight{where each element denotes the location or pose of a landmark (e.g., ultra-wideband tags, 3D points of visual features, and daily objects)}. In the SLAM problems, we consider two types of measurements: i) odometry and ii) landmark measurements (e.g., distances, directions, and relative poses). Let $\odomVals_t:=(\odomVal_1,\odomVal_2,\ldots,\odomVal_{t})$ be odometry measurements of which each is modeled by the likelihood function $p(\odomVal_i|\rbtVar_{i-1},\rbtVar_{i})$.  Without loss of generality, for the ease of notation in the method formulation, we assume the robot makes at most a landmark measurement at a time. Let $\measVals_t:=(\measVal_0,\measVal_1,\ldots, \measVal_t)$ be all landmark measurements in which each is modeled by the likelihood function $p(\measVal_i|\rbtVar_{i},\lmkVar_{d_i})$, where $\daVal_i \in \{1,2,\ldots,n\}$ is the landmark index associated with measurement $\measVal_i$. Let $\daVals_t:=(\daVal_0, \daVal_1, \ldots, \daVal_t)$ be data associations for all landmark measurements thus the posterior of robot and landmark variables at time step $t$ can be formulated by
\begin{align}
& p(\rbtVars_t, \lmkVars_n|\measVals_t,\odomVals_t, \daVals_t) \\
& \stackrel{\text{Bayes}}{\propto} \prod_{i=0}^{t} p(\measVal_i|\rbtVar_{i},\lmkVar_{d_i})\prod_{i=1}^{t} p(\odomVal_i|\rbtVar_{i-1},\rbtVar_{i})p(\rbtVars, \lmkVars),
\end{align}
where $p(\rbtVars, \lmkVars)$ denotes the prior.

Our goal is to infer marginal posteriors. Our method represents marginals either by samples or parametric models, depending on the uncertainty of the marginal. This hybrid density modeling aims to achieve a balance between computational efficiency and expressiveness. Our method is inspired by a typical observation about the posterior: oftentimes, only a few of the landmarks have very uncertain marginal posteriors, while marginals of most variables can be well described by Gaussian distributions. For example, a noisy range-only \emph{or} bearing-only measurement between a robot pose and a landmark can simulate either spherical or conic landmark distributions in 3D, while accumulated measurements may well constrain the landmark to a concentrated and (almost) unimodal distribution.

Specifically, we use particle filters to estimate very uncertain landmarks for the expressiveness of sample-based density representations. Meanwhile, we maintain a Gaussian approximation of the posterior for computation efficiency. Combining these two techniques leads to a win-win situation over time: i) the Gaussian approximation provides particle filters with a parametric model of the smoothed robot path, reducing the sampling complexity to landmarks, and ii) particle filters afford the Gaussian approximation statistically probable values of landmarks, which can be used to explicitly reset linearization points in nonlinear optimization solvers, thereby benefiting the pursuit of global optima. We describe the details of our method in the following sections.

\subsection{Gaussian Approximation}
We denote the Gaussian approximation (GA) of the posterior at time step $t$ as $g(\rbtVars_t, \lmkVars_n|\measVals_t,\odomVals_t, \daVals_t)$. We categorize all landmarks into a set of non-Gaussian landmarks $\lmkVar_{\uct_t}$ and a set of Gaussian landmarks $\lmkVar_{\gaus_t}= \lmkVars_n \setminus \lmkVar_{\uct_t}$, where $\uct_t \cup \gaus_t=\{1,2,\ldots,n\}$ and $\uct_t \cap \gaus_t=\emptyset$ (see Sec. \ref{sec:update-new-lmk-meas} for the approach to identify Gaussian landmarks). We use the Gaussian approximation to represent the posterior of robot poses and the Gaussian landmarks, resulting in
\begin{align}
& p(\rbtVars_t, \lmkVars_n|\measVals_t,\odomVals_t, \daVals_t)\\
& \stackrel{\text{GA}}{\approx} p(\lmkVar_{\uct_t}|\rbtVars_t, \lmkVar_{\gaus_t},\measVals_t,\odomVals_t, \daVals_t)g(\rbtVars_t,  \lmkVar_{\gaus_t}|\measVals_t,\odomVals_t, \daVals_t), \label{eqn:ga}\\
& \stackrel{\text{CIR}}{=} p(\lmkVar_{\uct_t}|\rbtVars_t, \measVals_t,\odomVals_t, \daVals_t)g(\rbtVars_t, \lmkVar_{\gaus_t}|\measVals_t,\odomVals_t, \daVals_t), \label{eqn:ci-reduction} \\
& \stackrel{\text{CIR}}{=} \prod_{i \in \uct_t} p(\lmkVar_i|\rbtVar_{\idx_i}, \measVal_{\idx_i},\daVal_{\idx_i})g(\rbtVars_t, \lmkVar_{\gaus_t}|\measVals_t,\odomVals_t, \daVals_t) \label{eqn:post-fac}
\end{align}
where $\rbtVar_{\idx_i}$ denotes the set of robot poses that observed landmark $\lmkVar_i$ (i.e., $\idx_i=\{j \in [0,t]| \daVal_j = i\}$ denotes the time steps observing landmark $\lmkVar_i$). The reduction from \eqref{eqn:ga} to \eqref{eqn:ci-reduction} exploits the conditional independence relation (CIR) that landmarks are conditionally independent given a robot path. Note that while we use the CIR in a similar way to typical Rao-Blackwellized particle filters (RBPFs) for SLAM (e.g., FastSLAM 2.0 \cite{montemerlo2003fastslam}), our probabilistic modeling and inference methods differ from these RBPFs. The Gaussian approximation can be computed via the Laplace approximation \cite{bishop2006gaussian}: 1) finding the MAP estimate as the mean and 2) estimating the inverse of the Hessian of the negative logarithm of the posterior at the MAP estimate as the covariance. In practice, many nonlinear local optimization solvers, such as GTSAM, are capable of providing efficient out-of-the-box solutions to the Gaussian approximation, although they are subject to local optima of the MAP estimate. In this paper, we refer to them as Gaussian solvers.
\begin{align}
& p(\rbtVars, \lmkVars|\measVals)\\
& \stackrel{\text{GA}}{\approx} p(\lmkVar_{\mathcal{N}}|\rbtVars, \lmkVar_{\mathcal{G}},\measVals)g(\rbtVars,  \lmkVar_{\mathcal{G}}|\measVals),\\
& \stackrel{\text{CIR}}{=} \prod_{i \in \mathcal{N}} p(\lmkVar_i|\rbtVar_{\idx_i},\measVals)g(\rbtVars, \lmkVar_{\mathcal{G}}|\measVals)
\end{align}

\subsection{Marginal Posterior of non-Gaussian Landmarks}
\label{sec:marginal-posterior}
We will use particles to represent the marginal posterior of any non-Gaussian landmark $\lmkVar_i \in \lmkVar_{\uct_t}$. Alg. \ref{algo:lmksample} summarizes how we draw the particles. Integrating out all variables except $\lmkVar_i$ in \eqref{eqn:post-fac}, we can formulate the marginal posterior by
\begin{align}
& p(\lmkVar_i|\measVals_t,\odomVals_t, \daVals_t)\\
& \stackrel{\text{GA}}{\approx} \int_{\rbtVar_{\idx_i}}  p(\lmkVar_i|\rbtVar_{\idx_i}, \measVal_{\idx_i},\daVal_{\idx_i})g(\rbtVar_{\idx_i}|\measVals_t,\odomVals_t, \daVals_t) \label{eqn:marginal}\\
& \stackrel{\text{MC}}{\approx} \frac{1}{K} \sum_{k=1}^{K} p(\lmkVar_i|\rbtVar_{\idx_i}=\rbtVal_{\idx_i}^{(k)}, \measVal_{\idx_i},\daVal_{\idx_i}), \label{eqn:MC}
\end{align}
where $\{\rbtVal_{\idx_i}^{(k)}\}_{k=1}^K$ are $K$ i.i.d. samples of part of the robot path drawn from the Gaussian marginal $g(\rbtVar_{\idx_i}|\measVals_t,\odomVals_t, \daVals_t)$ (line \ref{line:gs-samples} in Alg. \ref{algo:lmksample}). Monte Carlo (MC) integration is applied in \eqref{eqn:MC}. The conditional of $\lmkVar_i$ in \eqref{eqn:MC} is a result of eliminating $\lmkVar_i$ from factors adjacent to $\lmkVar_i$ (line \ref{line:lmk-factors} in Alg. \ref{algo:lmksample}), thus
\begin{align}
p(\lmkVar_i|\rbtVar_{\idx_i}, \measVal_{\idx_i},\daVal_{\idx_i}) =  \frac{\potential(\lmkVar_i, \rbtVar_{\idx_i})}{\int_{\lmkVar_i} \potential(\lmkVar_i, \rbtVar_{\idx_i}) },
\end{align}
where
\begin{align}
\potential(\lmkVar_i, \rbtVar_{\idx_i})=\prod_{j \in \idx_i} p(\measVal_j|\rbtVar_{j},\lmkVar_i)p(\lmkVar_i).\label{eq:lmk-potential}
\end{align}
We use importance sampling to draw samples representing $p(\lmkVar_i|\rbtVar_{\idx_i}=\rbtVal_{\idx_i}^{(k)}, \measVal_{\idx_i},\daVal_{\idx_i})$. We design the proposal distribution of landmark $\lmkVar_i$ as the sum-mixture of binary factors, as seen in
\begin{align}
q(\lmkVar_i; \rbtVal_{\idx_i}^{(k)}) = \frac{1}{|\idx_{i}|} \sum_{j \in \idx_i} p(\lmkVar_i|\rbtVar_{j} = \rbtVal_j^{(k)},\measVal_j). \label{eqn:proposal}
\end{align}
This proposal\footnote{Although we sacrificed sample diversity, we adopted a lighter-weight proposal in our experiments to ensure computational efficiency. Specifically, we limited the number of components in the sum-mixture to 5 and chose these components randomly.} was chosen for two reasons: i) covering more area in the space of the landmark variable and ii) it is efficient to draw landmark samples $\{\lmkVal_i^{(m,k)}\}_{m=1}^{M}$ from the proposal $q(\lmkVar_i)$ by sampling a multinomial distribution and binary factors independently. With proposal samples $\{\lmkVal_i^{(m,k)}\}_{m=1}^{M}$ in hand, the normalized weight of each sample can be computed by
\begin{align}
w^{(m,k)} \propto \frac{p(\lmkVal_i^{(m,k)}|\rbtVal_{\idx_i}^{(k)}, \measVal_{\idx_i},\daVal_{\idx_i})}{q(\lmkVal_i^{(m,k)}; \rbtVal_{\idx_i}^{(k)})} \propto \frac{\potential(\lmkVal_i^{(m,k)}, \rbtVal_{\idx_i}^{(k)})}{q(\lmkVal_i^{(m,k)}; \rbtVal_{\idx_i}^{(k)})},\label{eqn:weights}
\end{align}
where $\sum_{m=1}^M w^{(m,k)}= 1$ (line \ref{line:weights} in Alg. \ref{algo:lmksample}). Finally we apply the re-sampling and regularization steps in \cite[Alg. 6]{arulampalam2002tutorial} to generate equally weighted samples of $\lmkVar_i$ (line \ref{line:resampling} in Alg. \ref{algo:lmksample}).

\setlength\floatsep{2pt}
\begin{algorithm}[t]
  \fontsize{9pt}{9pt}\selectfont
  \DontPrintSemicolon 
  \KwIn{Landmark $l$, Gaussian solver $g$}
	Draw robot path samples $\mathcal{X}$ from the Gaussian approx. \label{line:gs-samples}
	
	$\mathcal{F}=\{f(l,\mathbf{x})\} \gets$ All factors adjacent to $l$ \label{line:lmk-factors} \tcp*[f]{robot vars $\mathbf{x}$} 

	$\psi(l,\mathbf{x}) \gets$ Product of all factors in $\mathcal{F}$ \tcp*[f]{potential func. \eqref{eq:lmk-potential}}

	$q(l,\mathbf{x}) \gets$ Sum of elements in $\mathcal{F}$ \tcp*[f]{proposal density \eqref{eqn:proposal}}

	Initialize a set $\mathcal{S}$ for storing landmark samples

	\For{\text{\normalfont sample $\hat{\mathbf{x}}$ in $\mathcal{X}$}}{
		Draw landmark samples $\mathcal{L}$ from density $q(l,\mathbf{x}=\hat{\mathbf{x}})$

		Weights $\mathcal{W}$ of landmark samples by computing $\psi/q$ \label{line:weights}

		$\mathcal{L} \gets$ Resampling using $\mathcal{L}$ and $\mathcal{W}$ \label{line:resampling}
		
		Add $\mathcal{L}$ to $\mathcal{S}$
	}
	\Return{$\mathcal{S}$} \tcp*[f]{samples of $l$}
  \caption{LandmarkSampler}
  \label{algo:lmksample}
\end{algorithm}
\setlength\floatsep{2pt}
\begin{algorithm}[!t]
  \fontsize{9pt}{9pt}\selectfont
  \DontPrintSemicolon 
  \KwIn{New factor $f$ of landmark $l$, Gaussian solver $g$, dictionary $\mathcal{D}$ of non-Gaussian landmark samples}    
	\If(\tcp*[f]{new landmark}){l \textup{not in the Gaussian solver}}
	{
		Add $l$ to non-Gaussian landmarks $\mathcal{D}$ as a new key \label{line:new-lmk}
	}

	Add $f$ to the Gaussian solver \label{line:add-to-gs}

	\If{l \textup{in non-Gaussian landmarks} $\mathcal{D}$}
	{
		$\mathsf{LandmarkReinitializer}$($l$,$\ g$,$\ \mathcal{D}[l]$) \label{line:re-init} \tcp*[f]{re-init. in the solver}
	}
	Update the MAP estimate in the Gaussian solver \label{line:update-gs}
		
	\If{l \textup{in non-Gaussian landmarks} $\mathcal{D}$}
	{
		$\mathcal{D}[l]=\mathsf{LandmarkSampler}$($l$,$\ g$) \label{line:update-samples} \tcp*[f]{update samples}

		Compute empirical covariance matrix $C$ of $\mathcal{D}[l]$

		\If{\textup{the largest eigenvalue of $C$ is small}}
		{
			Delete $l$ from $\mathcal{D}$ \label{line:unimodal-lmk} \tcp*[f]{no longer non-Gaussian lmks}
		}
	}
	
	\Return{$g$ \textup{and} $\mathcal{D}$} \tcp*[f]{updated solver and dictionary}
  \caption{GAPSLAM}
  \label{algo:gapslam}
\end{algorithm}

\setlength\floatsep{2pt}
\setlength\textfloatsep{5pt}
\begin{algorithm}[!t]
  \fontsize{9pt}{9pt}\selectfont
  \DontPrintSemicolon 
  \KwIn{Landmark $l$, Gaussian solver $g$, samples $\mathcal{S}$ of landmark $l$}
	$\psi(l,\mathbf{x}) \gets$ Product of all factors adjacent to $l$ \tcp*[f]{robot vars $\mathbf{x}$}

	$\hat{l},\hat{\mathbf{x}} \gets$ Current estimate in the Gaussian solver $g$

	$l^{*} \gets$ Value in $\{\mathcal{S}, \hat{l}\}$ maximizes function $\psi(l,\mathbf{x}=\hat{\mathbf{x}})$ \label{line:maximize-posterior}

	Reset the value of $l$ in the Gaussian solver to $l^{*}$ \label{line:reset-value}
	
	\Return{$g$} \tcp*[f]{updated solver}
  \caption{LandmarkReinitializer}
  \label{algo:reinit}
\end{algorithm}

Thus the marginal posterior in $\eqref{eqn:MC}$ can be approximated by samples, as seen in
\begin{align}
& p(\lmkVar_i|\measVals_t,\odomVals_t, \daVals_t) \stackrel{\text{MC}}{\approx} \tilde{p}(\lmkVar_i|\measVals_t,\odomVals_t, \daVals_t) \\
& = \frac{1}{K} \sum_{k=1}^{K} \sum_{m=1}^{M} \delta(\lmkVar_i-\lmkVal_i^{(m,k)}). \label{eqn:estimate-marginal}
\end{align}
To summarize, for each of the $K$ robot path samples, we draw $M$ landmark samples. The delta functions (or other kernels) of all $KM$ landmark samples can be used to approximate the marginal posterior of the landmark.

\subsection{Updates with new odometry and landmark measurements}\label{sec:update-new-lmk-meas}
The previous sections describe how the Gaussian approximation helps to compute the particle-based belief of landmarks. Now we discuss how these particles provide the Gaussian approximation with linearization points, thus leading to the uncertainty-aware re-initialization in the Gaussian approximation. We will begin by analyzing the incremental update by one time step.

From time step $t$ to time step $t+1$, the robot receives a new odometry reading $o_{t+1}$ and a new landmark measurement $z_{t+1}$. It is easy to create a new robot pose $\rbtVar_{t+1}$ and absorb the odometry $\odomVal_{t+1}$ to form the intermediate Gaussian $g(\rbtVars_{t+1}, \lmkVars_n|\measVals_t,\odomVals_{t+1}, \daVals_t)$. However, fusing the measurement $\measVal_{t+1}$ is a more involved process. Alg. \ref{algo:gapslam} summarizes steps for tackling the landmark measurement. Assuming that the measurement $z_{t+1}$ comes from a landmark with index $d_{t+1}$\footnote{Data association and creation of new landmarks are left to the specific application in Sec. \ref{sec:da}.}, several cases are considered:

\begin{enumerate}
\item If the observed landmark $\lmkVar_{d_{t+1}}$ is not new or a non-Gaussian landmark (lines \ref{line:add-to-gs} and \ref{line:update-gs} in Alg. \ref{algo:gapslam}), we simply add the likelihood model of $\measVal_{t+1}$ (i.e., factor in the algorithm) to the Gaussian solver to update the Gaussian approximation without making any further changes.

\item If the observed landmark $\lmkVar_{d_{t+1}}$ is new (line \ref{line:new-lmk} in Alg. \ref{algo:gapslam}), i.e. $d_{t+1}=n+1$, we update the landmark set to $\lmkVars_{n+1} = \lmkVars_n \cup \{\lmkVar_{n+1}\}$ and indices of non-Gaussian landmarks to ${\uct_{t+1}}={\uct_{t}}\cup \{{n+1}\}$. We then add the likelihood model of measurement $\measVal_{t+1}$ to the Gaussian solver. Finally, we can simulate equally-weighted samples of the landmark using this single measurement and randomly select a point from the samples as the initial value of the landmark.

\item If the observed landmark $\lmkVar_{d_{t+1}}$ is not new and is in non-Gaussian landmarks $\lmkVar_{\uct_t}$, this is the challenging case we will focus on.
\end{enumerate}

In the third case, our goal is to use the new measurement $\measVal_{t+1}$ to determine whether we should explicitly re-initialize the non-Gaussian landmark in the Gaussian solver (line \ref{line:re-init} in Alg. \ref{algo:gapslam}). Furthermore, if our belief of the landmark converges to a less uncertain situation after fusing measurement $\measVal_{t+1}$, we remove the landmark from the non-Gaussian set $\lmkVar_{\uct_t}$ (line \ref{line:unimodal-lmk} in Alg. \ref{algo:gapslam}).

Algorithm \ref{algo:reinit} summarizes how we re-initialize a non-Gaussian landmark $\lmkVar_i$. As described in Sec. \ref{sec:marginal-posterior} and Alg. \ref{algo:lmksample}, we already have samples of the landmark at hand, which can be passed to Alg. \ref{algo:reinit} as input. The union of the samples and the current estimate of the landmark forms a set of candidate points for re-initialization. The point in the set that maximizes the posterior will be used to re-initialize the landmark (lines \ref{line:maximize-posterior} and \ref{line:reset-value} in Alg. \ref{algo:reinit}). During the evaluation of the posterior, all variables other than the landmark are fixed by the current mean of the Gaussian approximation. This reduces the evaluation to only the product of factors that are adjacent to the landmark, simplifying the computation. The re-initialization can be formulated as

\begin{align}
l_i^* = \argmax_{\lmkVal_i \in \{\mathcal{S}, \hat{\lmkVal}_i \} } \potential(\lmkVar_i=\lmkVal_i, \rbtVar_{\idx_i}=\hat{\rbtVal}_{\idx_i}),\label{eq:reinit}
\end{align}
where the potential $\potential(\cdot)$ is the same as \eqref{eq:lmk-potential}, $\mathcal{S}=\{ l_i^{(j)}\}_{j=1}^{M}$ denotes current samples of the landmark, and $\hat{l}_i$ and $\hat{\rbtVal}_{\idx_i}$ denote current estimates of the landmark and robot poses in the Gaussian solver.

After performing the re-initialization, we compute the Gaussian approximation for time step $t+1$ (line \ref{line:update-gs} in Alg. \ref{algo:gapslam}), $g(\rbtVars_{t+1},\lmkVars_n|\measVals_{t+1},\odomVals_{t+1}, \daVals_{t+1})$, where the new measurement $\measVal_{t+1}$ has been incorporated. Given the new Gaussian approximation, we update samples of the marginal posterior of $\lmkVar_i$ (line \ref{line:update-samples} in Alg. \ref{algo:gapslam}), which are cached for potential re-initialization in the future. Following \cite{blanco2008pure}, we compute an empirical covariance matrix of these samples as well as its largest eigenvalue. When the largest eigenvalue falls below a threshold, it indicates that the belief of landmark $\lmkVar_i$ has become sufficiently certain, and $\lmkVar_i$ is removed from the non-Gaussian landmark set $\lmkVar_{\uct_{t+1}}$ (line \ref{line:unimodal-lmk} in Alg. \ref{algo:gapslam}). As a result, $\lmkVar_i$ will no longer be estimated by particle filters and will only be involved in updates in the Gaussian solver. \highlight{Note that currently, we set the threshold of the eigenvalue with predetermined values. One can explore automatic approaches such as monitoring the convergence rate of the largest eigenvalue or performing normality tests on samples}.

%% file: results.tex
\section{Experiments and Results}
We implemented Algorithms \ref{algo:lmksample}-\ref{algo:reinit} in Python by blending our in-house code of particle filters and the Gaussian solver provided by the GTSAM library. In addition, we applied explicit regularization to the Gaussian solver by adding large-covariance priors on new non-Gaussian landmarks. These prior factors were not involved in the sampling or re-initialization and were removed once the corresponding landmarks were removed from non-Gaussian landmarks. We perform two sets of experiments: 1) we validate our solutions for distribution and point estimations using range-only SLAM experiments, and 2) we demonstrate the generalizability and scalability of our method using object-based bearing-only SLAM experiments. All experiments were conducted on a laptop with a 2.30GHz Intel Core i7-10875H CPU running Ubuntu 20.04.4 LTS.

\subsection{Range-only SLAM} \label{sec:range-only-slam}
\subsubsection{Datasets and methods for comparison}
The performance of GAPSLAM is evaluated by comparing it with other methods on a range-only SLAM dataset, as range measurements can easily lead to highly non-Gaussian/multi-modal posteriors. We use the Plaza 1 dataset \cite{djugash2009navigating} which provides time-stamped range and odometry measurements collected by a mobile robot in a planar environment. Ranges between the robot and four landmarks were measured by an ultra-wide band ranging system so each range measurement was tagged with the identity of the landmark. Two methods were used for comparison: 1) NSFG \cite{huang2021reference}, which directly draws samples from the joint posterior using nested sampling methods \cite{skilling2006nested, speagle2020dynesty} and is considered as reference solutions at the expense of computational burden, and 2) RBPF-SOG \cite{blanco2008efficient}, which models landmark conditionals as sum-of-Gaussians (SOGs) in the RBPF framework, updates the SOGs using multi-hypothesis EKFs, and attains real-time operation. We present two sets of results: 1) marginal posteriors for demonstrating how the Gaussian approximation helps particle filters to draw landmark samples, and 2) point estimates for validating the use of particle filters in re-initializing the Gaussian approximation.

\subsubsection{Results on marginal posteriors}
\begin{figure}[t]
    \centering
    \includegraphics[width=1.0\linewidth]{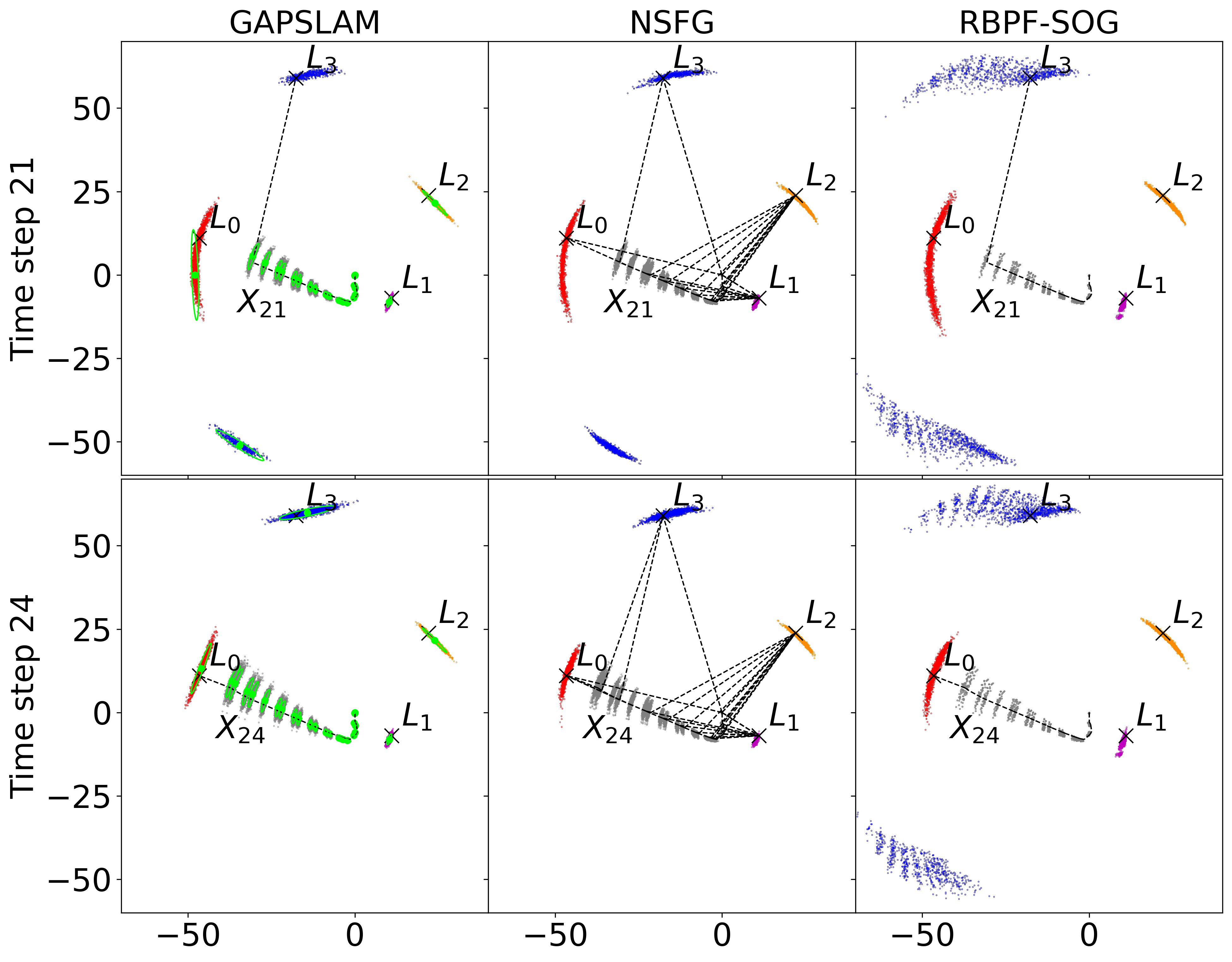}
    \caption{Samples from marginal posteriors by three methods at two time steps. The robot moves from $(0,0)$ to pose $X_{21}$ or $X_{24}$, observing 4 landmarks $L_{0-3}$. Black lines and markers indicate the ground truth. Measurements are shown as dashed lines (accumulated measurements are shown in the middle column). Gray dots indicate robot position samples while colored dots denote landmark samples. Green ellipses in GAPSLAM represent confidence intervals (2$\sigma$) of the Gaussian approximation.}
    \label{fig:ro-samples}
\end{figure}

\begin{figure*}[t]
    \centering
    \includegraphics[width=1.0\linewidth]{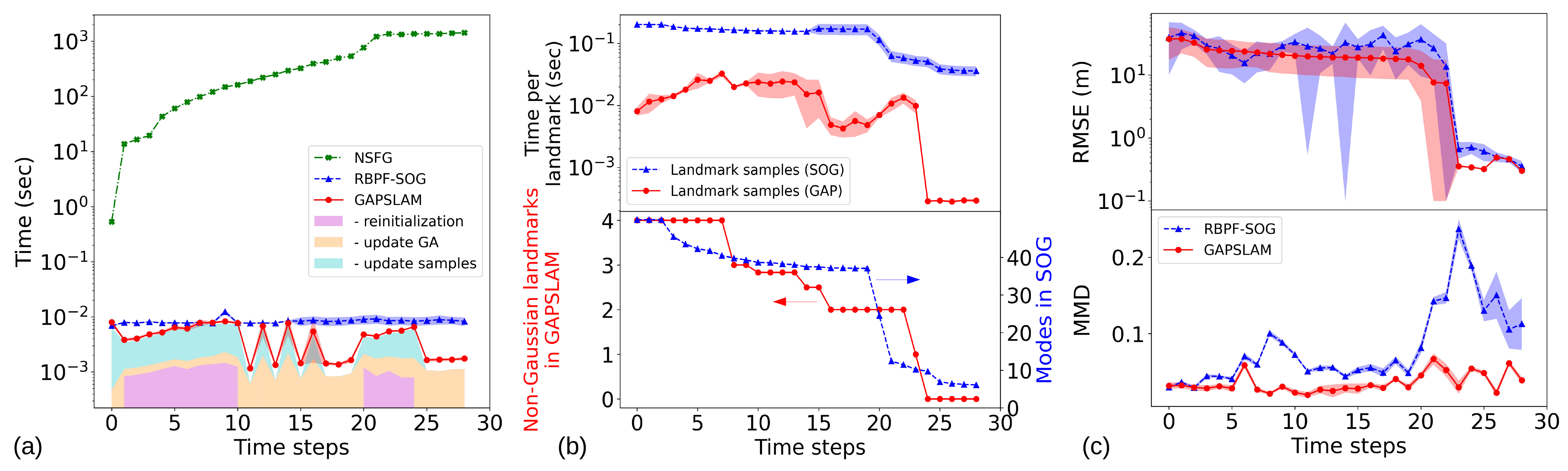}
    \caption{Performance comparison at early time steps in the Plaza1: (a) computational time for updating probabilistic models, (b) computational time for posterior samples of landmarks, the number of non-Gaussian landmarks in GAPSLAM, and the averaged number of modes per landmark in RBPF-SOG, and (c) the root mean square error (RMSE) of point estimates and the maximum mean discrepancy (MMD) \cite{gretton2012kernel} of landmark samples. The MMD is a distance between two distributions represented by samples. We compute the MMD between the reference samples and samples by GAPSLAM (or RBPF-SOG), so a lower MMD means samples that describe the posterior better. Error bands indicate the mean and standard deviation across 6 runs with different random seeds. }
    \label{fig:ro-perform}
\vspace{-5pt}
\end{figure*}

We present the posteriors of early time steps, which are expected to be strongly non-Gaussian (e.g., multi-modal). Fig.~\ref{fig:ro-samples} shows samples drawn by different methods to represent the posteriors. GAPSLAM shows good consistency with the reference samples by NSFG, while RBPF-SOG overestimates the uncertainty in the posteriors, as indicated by spurious modes of landmark $L_3$ samples (blue dots in the last column of Fig. \ref{fig:ro-samples}). The excessive number of modes is an expected result of tracking multiple hypotheses in SOGs. Although we applied the strategy suggested in \cite{blanco2008efficient} to prune hypotheses with negligible weights, as shown by the decreasing blue line in the bottom of Fig. \ref{fig:ro-perform}b, the averaged number of modes per landmark is still 11 at time step 21.

\setlength\textfloatsep{10pt}

In order to explain how GAPSLAM draws the samples in Fig. \ref{fig:ro-samples} and when the re-initialization occurs, we can start by looking at the green confidence intervals, which represent the Gaussian approximation. Samples of robot positions and Gaussian landmarks are drawn from the Gaussian. At time step 21, $L_0$ and $L_3$ are non-Gaussian landmarks, and we use the method described in Sec. \ref{sec:marginal-posterior} and Algorithm \ref{algo:lmksample} to draw their samples. Following \eqref{eq:reinit}, these samples are then used for re-initialization across time steps 21 and 24, which is reflected by the increased computation time for re-initialization in Fig. \ref{fig:ro-perform}a. Because the set of Gaussian landmarks extends from $(L_1, L_2)$ at time step 21 to all landmarks at time step 24, the number of non-Gaussian landmarks drops to zero, as shown in the bottom of Fig. \ref{fig:ro-perform}b.

Fig. \ref{fig:ro-perform}a illustrates the computation time of different methods for updating their probabilistic models. Note that the computation time for drawing dense samples ($K=200,M=100$ in \eqref{eqn:estimate-marginal}) of landmarks shown in Fig. \ref{fig:ro-samples} is separately shown in the top of Fig. \ref{fig:ro-perform}b. There are two reasons for excluding it from Fig. \ref{fig:ro-perform}a: 1) sampling landmark posteriors can be implemented as a parallel process, and 2) RBPF-SOG does not need landmark samples to update the SOGs, and GAPSLAM only needs sparse samples to assist the re-initialization to the Gaussian approximation. For the sparse samples (i.e., `update samples' in Fig. \ref{fig:ro-perform}a), we draw landmark samples given only the current mean of the robot path (i.e., $K=1$ in \eqref{eqn:estimate-marginal}). While sampling landmark posteriors consumes slightly more computation time than updating models in GAPSLAM before time step 24, the sampling still supports an update frequency of at least 30 Hz. After time step 24, there are no non-Gaussian landmarks, so GAPSLAM is just the Gaussian solver with additional functions for sampling Gaussian marginals, which enjoys significantly faster speeds.

Fig. \ref{fig:ro-perform}c presents the performance evaluation of point estimates and distribution estimation using the root mean square error (RMSE) and maximum mean discrepancy (MMD) \cite{gretton2012kernel}, respectively. We compute the MMD between samples by GAPSLAM (or RBPF-SOG) and the reference samples by NSFG. The MMD is a distance between two distributions represented by samples, so a lower MMD here indicates a more accurate set of samples for representing posteriors. The results show that GAPSLAM significantly outperforms RBPF-SOG in terms of MMD, especially after time step 21 when the spurious modes in SOGs become more prominent, providing quantitative evidence of the superior accuracy of GAPSLAM in sampling posteriors. Additionally, it is worth noting that the RMSE of GAPSLAM only slightly outperforms that of RBPF-SOG at these early time steps. \highlight{This indicates that MMD is a more suitable evaluation metric than RMSE for \emph{full posterior inference}, especially in highly non-Gaussian settings}.

\subsubsection{Results on point estimates}
Table \ref{table:rmse-recall} presents the errors of point estimates for the entire Plaza1 sequence. RBPF-SOG incurs less accurate and consistent results due to the particle depletion issue. GAPSLAM without re-initialization is essentially the Gaussian solver with additional priors on landmarks. However, only 58$\%$ of runs without re-initialization return solutions despite having RMSE values comparable to those of GAPSLAM. The superior performance of GAPSLAM supports the effectiveness of the sample-based, uncertainty-aware re-initialization.

\begin{table}[t]
\vspace{+0.3cm}
\caption{Point estimates for the full sequence of the Plaza1 dataset. The RMSE is presented by the mean $\pm$ standard deviation across 50 runs with different random seeds \highlight{which lead to different random initial values in the Gaussian solver}. The case `GAP. w/o reinit.' serves as an ablation study and is configured by disabling the non-Gaussian landmarks set and operations for re-initialization and sampling, which is essentially just GTSAM with the explicit regularization treatment (large-covariance priors). The RMSE in this case is computed on 29 out of 50 runs (i.e., $58\%$ success rate) that were not ceased by \highlight{the GTSAM error, indeterminant linear system. The GTSAM error is expected because, in this problem, bad initial values may incur near-singular linear systems}.}
\label{table:rmse-recall}
\begin{tabularx}{\linewidth}{@{} p{1.9cm} p{0.5cm} l l l @{}}
\toprule
Metric  
& Odom. & GAPSLAM & RBPF-SOG & GAP. w/o reinit.\\ 
\midrule
RMSE (cm) & 639.6   & $\mathbf{34.4\pm 0.0}$ & $56.0\pm 5.4$ & $34.5\pm 0.0$ \\ 
Success rate (\%) &- & $\mathbf{100}$     & $\mathbf{100}$        & 58  \\ 
\bottomrule
\end{tabularx}
\vspace{-5pt}
\end{table}

\subsection{Object-based bearing-only SLAM} \label{sec:obj-slam}
\subsubsection{Datasets and real-world experiments} We estimate 6DOF camera poses and a map of object locations, using visual odometry and object detections from RGB videos. We aim to demonstrate the scalability of GAPSLAM to three-dimensional (3D) environments and its ability to fuse other types of measurements such as bearing-only. We test GAPSLAM using RGB data from the RGB-D Scenes Dataset v2 \cite{lai2014unsupervised} and a video collected by a monocular camera in our office area. For visual odometry, we use camera poses of key frames computed by ORB-SLAM3 \cite{ORBSLAM3_TRO}, while object classes and masks in the key frames are detected using the Detic detector \cite{zhou2022detecting}. \highlight{Camera poses are optimized using both visual odometry and measurements to objects.}

\begin{figure*}[t]
    \centering
    \includegraphics[width=1.0\linewidth]{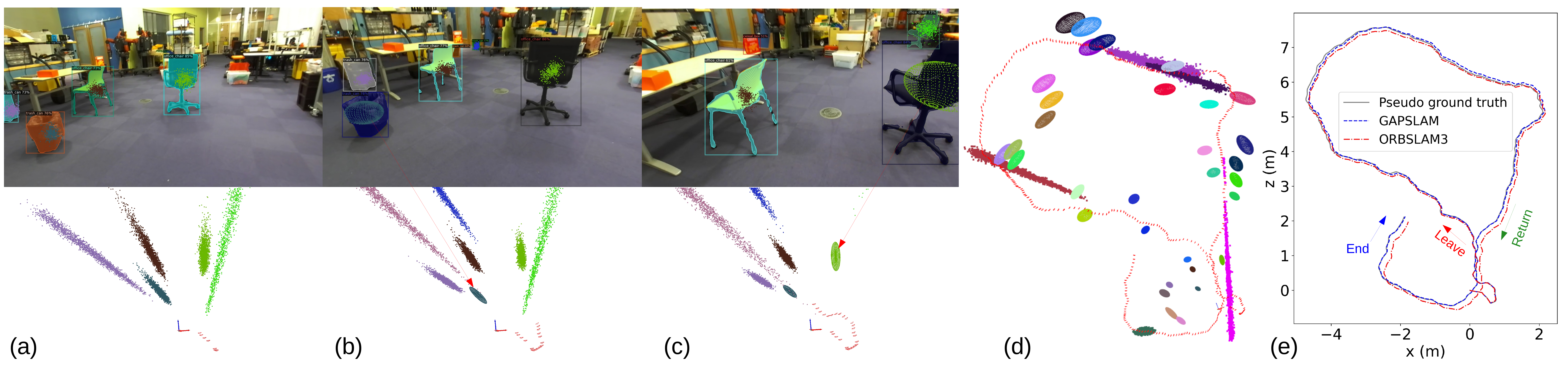}
    \caption{Estimates made by GAPSLAM for the realworld RGB sequence: (a,b,c) samples or confidence intervals (3$\sigma$) of object locations in 3D and their reprojection in images at key frames 10, 20, and 30, (d) final estimates of camera poses and samples and confidence intervals of object locations (we show samples of only 5 non-Gaussian landmarks to avoid clutter), and (e) a comparison of trajectory estimates with the pseudo ground truth. ORBSLAM3 did not close the big loop between `leave' and `return' due to the large change ($\sim$180 degrees) in viewing angles.}
    \label{fig:obj-realworld}
\vspace{-5pt}
\end{figure*}

\subsubsection{Pre-processing object detections} We treat the center $z_k$ of an object mask on an image as a projection factor that models the reprojection error $h(x_i,l_j;K)-z_i$, where $h(\cdot)$ denotes the reprojection of 3D object center $l_j$ in the image with pose $x_i$ and camera intrinsics $K$. This measurement $z_k$ is either associated with an existing landmark or yields a new landmark, depending on the semantic and geometric information of the map. Thus wrong semantic labels may affect data association or spawn spurious landmarks. On the other hand, partial views of objects, which are often caused by occlusions or objects on image edges, lead to 2D mask centers that significantly deviate from the object center in 3D. To address these issues, we pre-process object detections as follows:
\begin{itemize}
\item Rejecting object detections that are potentially occluded by other detections on the image. Specifically, for any pair of detections (say $i$ and $j$) whose bounding boxes intersect, one of the detections (say $i$) will be discarded if any of the following conditions is satisfied:
\begin{itemize}
\item The object mask of detection $i$ is smaller than a fraction (e.g., $10\%$ in our experiments) of the mask of detection $j$. This condition is for rejecting very small masks.
\item The area of mask $i$ within the bounding-box intersection is smaller than that of mask $j$ and the area ratio between the intersection and the mask $i$ is greater than a fraction (e.g., $10\%$ in our experiments). This condition is for rejecting object detections that may be occluded by large bounding-box intersections.
\end{itemize}

\item Increasing uncertainty in noise models of object mask centers that are close to image edges. Let us denote the relative position of the center of an object mask on an image as $(r_x,r_y) \in [-1,1]^2$ so $r_x=0$ and $r_y=0$ indicate a mask center located at the image center and $|r_x|$ or $|r_y|$ that approaches 1 indicates a mask center that is close to image edges. Thus we define standard deviations in the noise model as
\begin{align}
\sigma_{x} &= \delta_{x}(1-|r_x|) + \text{max}(\delta_{x}, \delta_{y}) |r_x|,\label{eqn:noise}\\
\sigma_{y} &= \delta_{y}(1-|r_y|) + \text{max}(\delta_{x}, \delta_{y}) |r_y| \label{eqn:noise2}
\end{align}
where $\delta_{x}$ and $\delta_{y}$ denote standard deviations of the mask in pixels.

\item Rejecting any object detection which cannot be associated with landmarks in the same class via the maximum likelihood (ML) data association \cite{kaess2009covariance, neira2001data} but passes the data association with landmarks that belong to different classes. This step intends discarding detections with wrong object classes. See the following section about the data association.
\end{itemize}

\subsubsection{Data association and creation of new landmarks}\label{sec:da}
We choose to implement the ML data association with some additional treatment for exploiting semantic information in object detections and posterior samples in GAPSLAM. For a mask center $z_k$ labeled by an object class and observed from pose $x_i$, we first compute its Mahalanobis distances to reprojections of any landmark $l_j$ in the same class, as defined in 
\begin{align}
D^{2}_{kj} = \lVert h(\hat{\mathbf{x}})-z_k \rVert^2_{C}
\end{align}
where $\mathbf{x}=(x_i, l_j)$ and
\begin{align}
C = \frac{\partial h}{\partial \mathbf{x}} \bigg|_{\hat{\mathbf{x}}} \Sigma \frac{\partial h}{\partial \mathbf{x}} \bigg|_{\hat{\mathbf{x}}}^{T} + \Gamma.
\end{align}
$\Sigma$ is the covariance between $x_i$ and $l_j$ in the Gaussian approximation of the posterior. $\Gamma$ is the covariance of measurement noise model (e.g., \eqref{eqn:noise} and \eqref{eqn:noise2}). The measurement is associated with $l_j$ if $D^{2}_{kj}$ is the smallest among landmarks in the same class and passes the chi-square test, $D^{2}_{kj} < \mathcal{X}^2_{d,\alpha}$, where $d$ is the dimension of the measurement and $\alpha$ is the desired confidence level.

Note that the quality of the Gaussian approximation affects the ML data association so, if the ML data association does not accept any landmark, we then perform another round of data association using landmark samples. For any landmark $l_j$ in the same class as detection $z_k$, we reproject samples of $l_j$ onto the image and count the percent $p_{kj}$ of samples that fall in the bounding box of the detection. The detection is associated with $l_j$ if $p_{kj}$ is the highest among landmarks in the same class and higher than a threshold (e.g., $10\%$ in our experiments).

If the sample-based data association does not accept any landmark either, we create a new landmark in the map.

\subsubsection{Results}
Fig. \ref{fig:obj-realworld} shows qualitative results on our realworld sequence. We set the Detic detector to recognize only a certain number of classes: cup, cereal box, trash can, skateboard, office chair, football, bottle, traffic cone, and toy car since they are quite common in our office and we found Detic worked with good recall and precision for these classes. Figs. \ref{fig:obj-realworld}a, b, and c show how marginal posteriors of object locations evolve over key frames 10, 20, and 30, which occur before the `leave' arrow on the path in Fig. \ref{fig:obj-realworld}e. The trash can in Fig. \ref{fig:obj-realworld}b and the chair in Fig. \ref{fig:obj-realworld}c have been identified as Gaussian landmarks by GAPSLAM so we draw confidence intervals (3$\sigma$) of their locations using the Gaussian approximation in lieu of samples.

There are 415 camera poses and 100 landmarks being created in the end. Fig. \ref{fig:obj-realworld}d visualizes all camera poses and landmarks marginals in 3D. Fig. \ref{fig:obj-realworld}e compares the pseudo ground truth and the estimated trajectories by GAPSLAM and ORBSLAM3. ORBSLAM3 fails to close the big loop between `leave' and `return' since the camera returns from an opposite viewing angle, and the bag-of-word approach in ORBSLAM3 does not recognize the return. The trajectory by GAPSLAM is aligned with the pseudo ground truth much better since object detections introduce many loop closures along the path. We found object detections by Detic performed consistently well under big changes of viewing angles. In GAPSLAM, the big loop between `leave' and `return' is successfully closed when the camera returns to and recognizes chairs and trash cans that appeared at early key frames (Figs. \ref{fig:obj-realworld}a, b, and c). The pseudo ground truth is the early portion of ORBSLAM3 results on a longer video, which circles our office three times along the same path, so the early portion has been corrected by loop closures introduced in the later portion.

Table~\ref{table-object-rmse} shows the RMSEs of estimated trajectories on the Scenes v2 sequences and our realworld sequence. GAPSLAM attains lowest errors in all the sequences. \highlight{We attribute this accuracy to two reasons: 1) extra constraints introduced by object-based measurements, and 2) functions in the GAPSLAM algorithm.} We disable some functions in the GAPSLAM algorithm to demonstrate their contributions to the estimation of the path. We find the pre-processing makes the biggest impact to the estimation while the re-initialization ranks the second and the sample-based data association ranks the third. Disabling all of them incurs significant estimation errors in some of the sequences (e.g., seqs. 9 and realwrold).

\begin{table}[t]
\vspace{+0.5cm}
\caption{Root mean square error (cm) of estimated paths under different settings. Each row below GAPSLAM serves as an ablation study that disables one (or three) of the functions in the object-based SLAM system.}
\label{table-object-rmse}
\begin{tabularx}{\linewidth}{@{} l *{6}{L} l @{}}
\toprule
\multirow{2}{*}{Method} & \multicolumn{6}{l}{Sequences in RGB-D Scenes Dataset v2} & \multirow{2}{*}{\stackanchor{Real}{world}}\\ \cline{2-7} 
                        & 1   & 2   & 3   & 4      & 9   & 10   &\\ \midrule
ORBSLAM3                & 1.9 & 2.2 & 2.0 & 2.0    & 2.6 & 3.7   & 14.0\\
GAPSLAM                 & \bf{0.9} & \bf{1.0} & \bf{0.7} & \bf{1.2}    & \bf{0.7} & \bf{2.6}     & \textbf{5.2}\\
- no pre-processing       & 1.0 & \bf{1.0} & 0.8 & 1.4    & 2.4 & 2.7      & 6.6\\
- no sample-based DA    & \bf{0.9} & \bf{1.0} & \bf{0.7} & \bf{1.2}    & 0.8 & \bf{2.6}      & 5.8\\
- no reinitialization            & \bf{0.9} & \bf{1.0} & 0.8 & 1.3    & \bf{0.7} & \bf{2.6}     & 6.6\\
- disable all above     & 1.0 & 2.2 & 0.9 & 1.3    & 4.2 & 2.7      & 9.5\\
\bottomrule
\end{tabularx}
\end{table}
\setlength\floatsep{15pt}
\begin{table}[t]
\caption{Runtime profiling of the object-based SLAM system and specifics for different datasets.}
\label{table-object-time}
\begin{tabularx}{\linewidth}{@{} l *{6}{L} l @{}}
\toprule
\multirow{2}{*}{Items} & \multicolumn{6}{l}{Sequences in RGB-D Scenes Dataset v2} & \multirow{2}{*}{\stackanchor{Real}{world}}\\ \cline{2-7} 
                            & 1    & 2    & 3    & 4    & 9    & 10   &\\ \midrule
Time/frame (ms)              & 43.2 & 28.6 & 29.7 & 32.1 & 23.9 & 19.7 & 63.3\\
- pre-processing            & 4.2  & 2.7  & 2.2  & 3.0  & 1.5  & 1.1  & 9.0\\
- data association 		    & 6.2  & 4.6  & 7.6  & 7.2  & 7.0  & 4.8  & 9.8\\
- GAPSLAM                   & 32.8 & 21.3 & 21.2 & 21.9 & 15.4 & 13.9  & 44.4\\
\hspace{5pt} - reinit.      & 3.2  & 2.4  & 1.6  & 1.4  & 1.0  & 1.3  & 6.1\\
\hspace{5pt} - GA           & 8.8  & 6.1  & 7.6  & 8.8  & 4.8  & 3.7  & 13.9\\
\hspace{5pt} - samples      & 13.6 & 8.0  & 6.8  & 6.2  & 4.7  & 5.5  & 17.1\\ \midrule
Obj. det./frame (-)   		& 5.1  & 5.0  & 5.2  & 5.0  & 2.8  & 3.0  & 5.3\\
Key frames (-)      		& 99   & 95   & 108  & 109  & 96   & 87  & 415\\
$\#$ of landmarks (-)      	& 8    & 7    & 8    & 8    & 4    & 3   & 100\\
\bottomrule
\end{tabularx}
\end{table}

\begin{figure}[t!]
    \centering
    \includegraphics[width=1.0\linewidth]{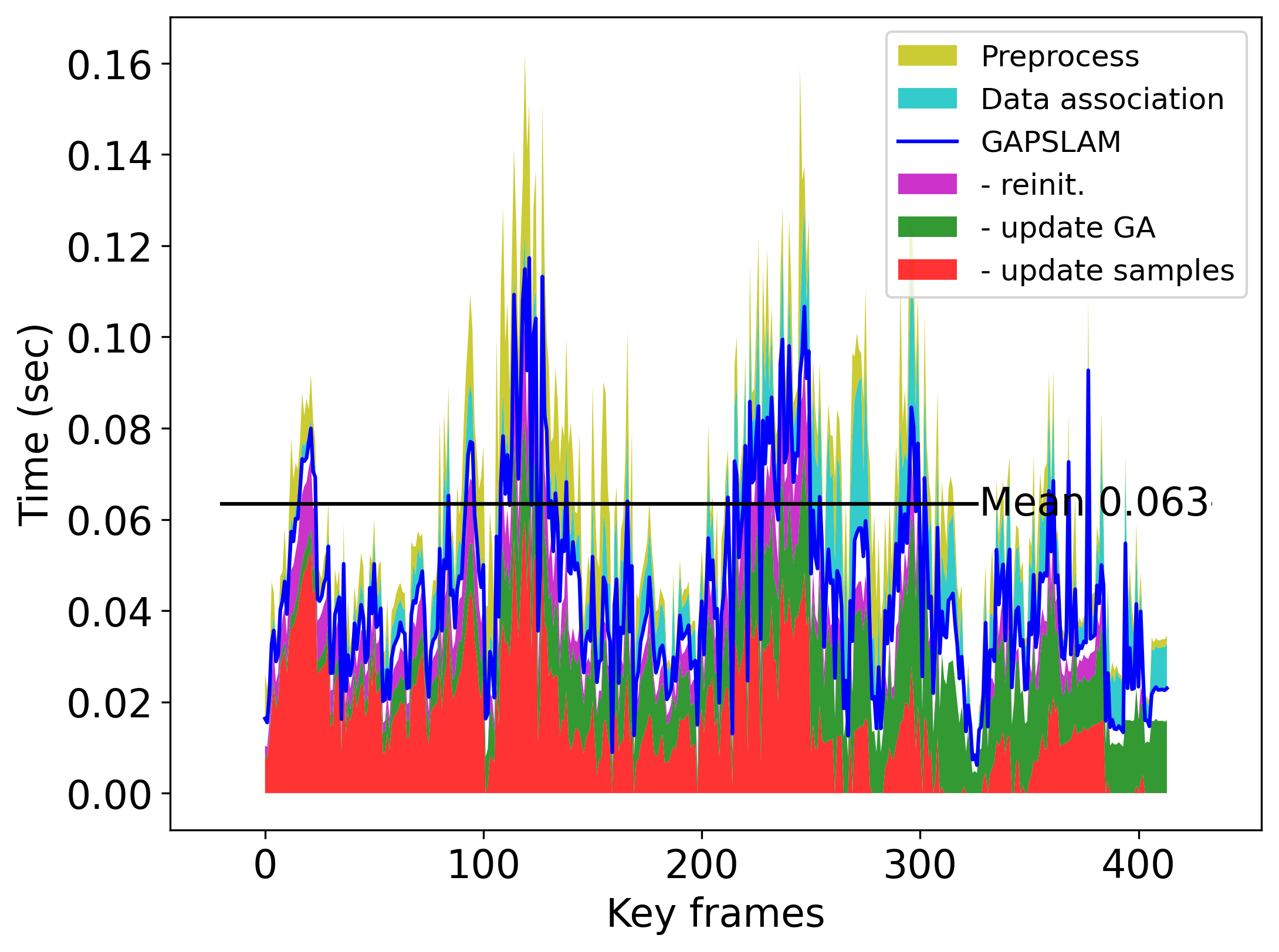}
    \caption{Runtime profiling of the object-based SLAM system (in Python).}
    \label{fig:obj-runtime}
\end{figure}

Fig. \ref{fig:obj-runtime} shows the runtime of our system for the realworld sequence. The system implemented in Python supports an update frequency of 6 Hz for the slowest key frames and, on average, affords an update frequency of 16 Hz. We are confident that the runtime can be greatly optimized by an improved implementation. Given that we only process key frames, GAPSLAM is able to support real-time operation. Table~\ref{table-object-time} shows the runtime profiling of our system and specifics of different datasets. The runtime on Scenes v2 sequences is expected to be faster since the sequences involve fewer poses and objects. The crux of averaged runtime is on steps for updating samples and the Gaussian approximation. However, Fig. \ref{fig:obj-runtime} shows that updating samples only dominates the runtime in early key frames, when many non-Gaussian landmarks had just been created. Revisiting objects at later key frames reduces the uncertainty in the posteriors of object locations, so more and more objects are moved to the set of Gaussian landmarks. Thus the computation burden gradually transfers to updating the Gaussian approximation. The scalability of updating the Gaussian approximation can be improved via many strategies such as fix-lag smoothing \cite{dellaert2017factor}.

%% file: conclusion.tex
\section{Conclusion}
We presented a real-time algorithm, GAPSLAM, that precisely infers non-Gaussian/multi-modal marginal posteriors encountered in SLAM. Our experiments justified the efficacy of the adaptive modeling strategy for efficient full posterior inference, and the uncertainty-aware re-initialization technique for explicitly correcting linearization points in nonlinear optimization solvers. Future work includes inferring the joint posterior of multiple variables, \highlight{incorporating more complex likelihood of measurements (e.g., learned orientation distributions~\cite{pmlr-v139-murphy21a}, multi-modal data association~\cite{doherty2019multimodal})}, and applying GAPSLAM to realworld planning tasks that tackle the non-Gaussian belief of obstacle locations \cite{han2022non}.